\documentclass{gretsi}
                    
\usepackage[english]{babel}   
\usepackage{times}			
\usepackage{graphicx}
\usepackage{caption}
\usepackage{makecell}
\usepackage{algorithm}
\usepackage{xcolor}
\usepackage{amsmath}
\usepackage{amssymb}
\usepackage[noend]{algpseudocode}
\usepackage{caption}
\usepackage{subcaption}
\usepackage{multirow}
\usepackage{multicol}
\usepackage{psfrag}
\usepackage[hidelinks]{hyperref}
\usepackage{stfloats}
\usepackage[utf8]{inputenc}
\usepackage[T1]{fontenc}
\usepackage{hyperref}
\usepackage{hhline}
\newlength{\abc}
\AtBeginDocument{%
\renewcommand{\ref}[1]{\mbox{\autoref{#1}}}

}

\newcommand{\setfont}[2]{{\fontfamily{#1}\selectfont #2}}
\newcommand{\etal}{ \textit{et al.} }

\title{Toward more frugal models for functional cerebral networks \\ automatic recognition with resting state fMRI}

\author{\coord{Lukman E. }{Ismaila}{1},
        \coord{Pejman}{Rasti}{1,2},
    \coord{Jean-Michel}{Lemée}{3},
    \coord{David}{Rousseau}{1*}}

\address{\affil{1}{LARIS, UMR INRAe, IRHS Angers,
         Universit\'e  d'Angers, France}
         \affil{2}{CERADE 
         ESAIP, École d'Ingénieurs}
         \affil{3}{Service de Neurochirurgie 
         CHU d’Angers, France}}
 
\email{david.rousseau@univ-angers.fr}

\frenchabstract{Nous considérons une situation d'apprentissage machine où des modèles à base de réseaux de neurones convolutionnels classiques ont montré de bonnes performance. Nous investiguons différentes techniques d'encodage sous forme de supervoxel puis de graphes pour réduire la complexité du modèle tout en limitant la perte de performance. Cette approche est illustrée sur une tâche de reconnaissance de réseaux fonctionnels de repos pour des patients atteints de tumeurs cérébrales. Les graphes encodant des supervoxels préservent les caractéristiques d'activation des réseaux cérébraux fonctionnels à partir des images, réduisant les paramètres du modèle de 26 fois tout en maintenant les performances du modèle CNN.}

\englishabstract{We refer to a machine learning situation where models based on classical convolutional neural networks have shown good performance. We are investigating different encoding techniques in the form of supervoxels, then graphs to reduce the complexity of the model while tracking the loss of performance. This approach is illustrated on a recognition task of resting-state functional networks for patients with brain tumors. Graphs encoding supervoxels preserve activation characteristics of functional brain networks from images, optimize model parameters by 26 times while maintaining CNN model performance.}

\begin{document}
\maketitle
 
\section{Introduction} 


Convolutional neural networks (CNN) are powerful tools to perform computer vision tasks. CNN are however very demanding in terms of energy, data and annotation due to the large amount of parameters to be tuned during their training. These limitations are specially important in medical imaging where the constitution of large cohorts of unhealthy patients can be a bottleneck as frequently observed in cases of rare diseases like brain tumor. Recently, we have shown the possibility to circumvent this limitation by the use of transfer learning from self-supervised training on healthy data to unhealthy data \cite{ismaila2022self}. We used small data in our experiments, and  approach opens the possibility for scalability when a larger model is trained from additional data acquired.

This was obtained for the automatic recognition of functional cerebral networks via resting-state functional magnetic resonance imaging (rs-fMRI) \cite{ismael2022deep} for patient with brain tumors. 
The CNN architecture proposed for the classification of functional brain network with 3D fMRI images by Ismaila\etal, was observed with high model training parameters despite the small data size \cite{ismael2022deep} which constitutes a complex model and struggles with risks of overfitting. 

\begin{figure}[!ht]
\includegraphics[scale=.062]{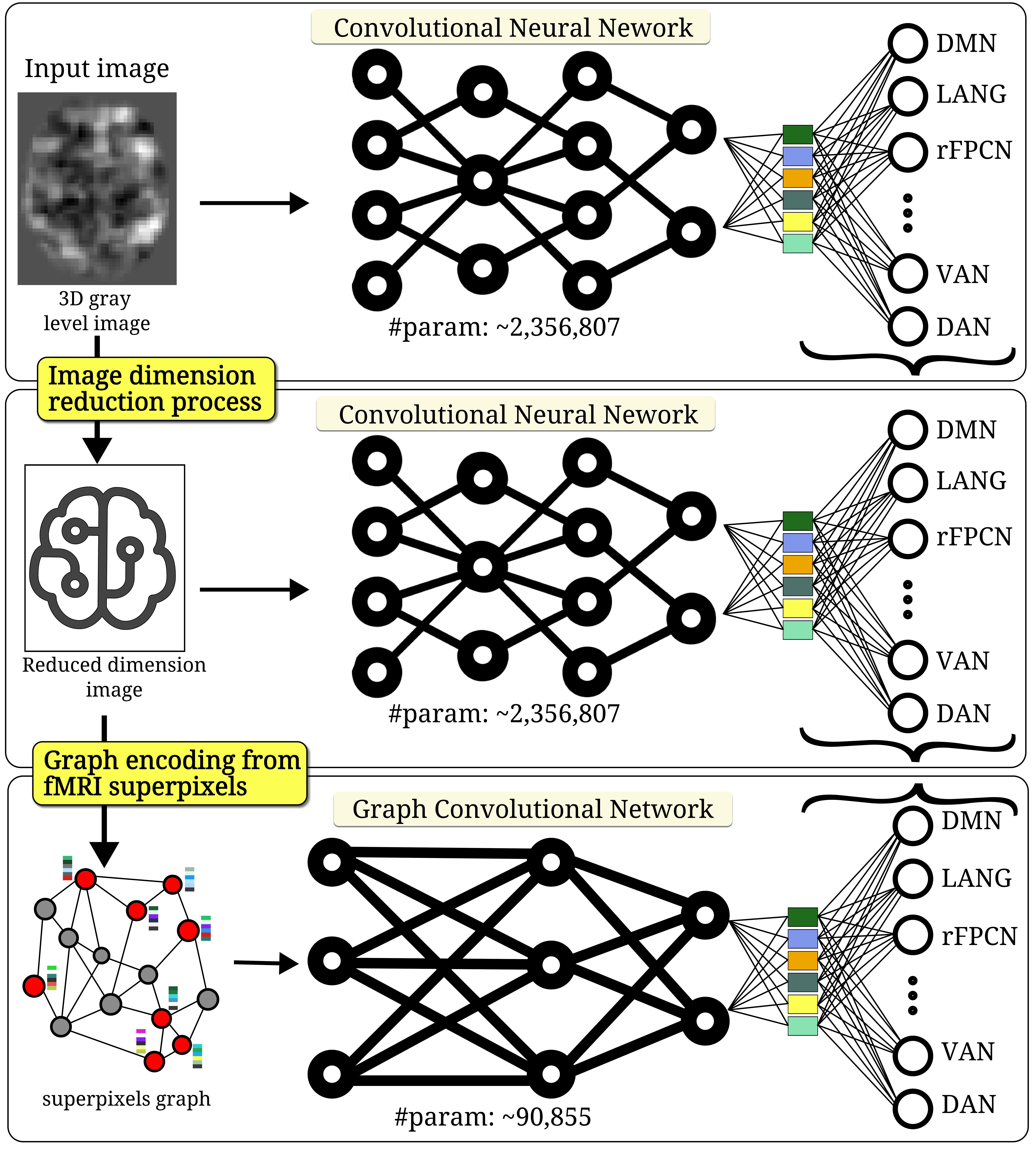} 
\caption{Visual abstract of our method. We consider as baseline performance either in terms of number of parameters and accuracy of CNN applied on raw rs-fMRI activation maps for functional cerebral network automatic recognition. We compare this performance with neural networks applied on compacted versions of the images.} 
\label{fig:gnn_abst}
\end{figure}

In this work, we test possible ways to simplify deep learning models by reducing the overall parameter size. To this purpose, we propose to compare a basic CNN method with the approach depicted in \autoref{fig:gnn_abst}. \color{black}{Based on a recent work by Gousia \etal, which highlighted the benefits of graph encoding in optimizing CNN model parameters especially in medical imaging \cite{habib2022optimization}. We investigate various ways of encoding the rs-fMRI 3D volume data in more compacted fashions and systematically compare our observation with the performance obtained in \cite{ismael2022deep}. This effort only represent an initial attempt towards more efficient encoding of our brain volume images, as well as opens the possibility for scalability when a larger model is trained from additional data acquired.}\color{black}

\section{Database}
 fMRI brain network activation image data of 81 healthy subjects and 55 unhealthy patients were collected. Regular volunteers provide the healthy data, while patients with brain tumors where a binary mask indicate region of lesion in the brain constitute the unhealthy data. This analysis, was done in separate components which creates brain maps of the regions with synchronous blood oxygen level dependent (BOLD) signal activity. In the data acquisition stage, we extracted the intrinsic connectivity networks (ICNs) by using methods that combine the information of both the temporal and spatial dimensions, such as independent component analysis. The extracted signals represent the neuro-anatomical basis for the functional networks in the brain \cite{lemee2019resting}.
 \color{black}{} The statistical parametric mapping (SPM) anatomy toolbox for Matlab was used to generate the 3D brain volume images, from the initial spatio-temporal fMRI signals.\color{black}{} Among the 55 ICNs processed for each patients, 7 of these signals where recognized manually by experts to be biological networks of the brain such as Default Mode Network (DMN), Language Network (LANG), Right Fronto-parietal Control Network (rFPCN), Left Fronto-parietal Control (lFPCN), Salience Network (SAL), Dorsal Attention Network (DAN) and Ventral Attention Network (VAN). The annotated images were used in two versions: full images (connectivity map) and corresponding thresholded images.

\section{Spatial dimension reduction}

\begin{figure}[!ht]
\centering
\includegraphics[scale=.048]{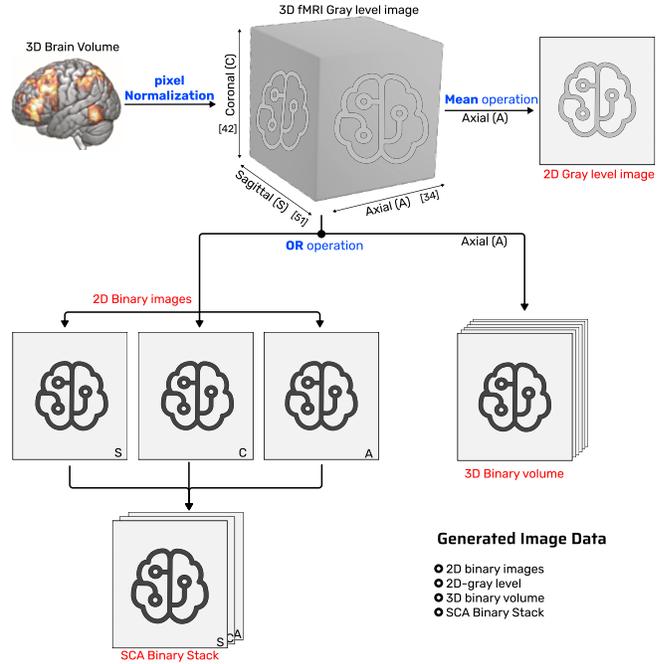}
\caption{fMRI image dimension reduction process.}
\label{fig:img_dm_reduction}
\end{figure}

One may wonder if the entire 3D volume in gray levels is fully informative for automatic recognition of the functional cerebral networks. Several dimension reduction approaches can be envisioned. From the acquired brain volumes of resting-state fMRI images $42 px\times51 px\times34 channels$, we normalized the pixel intensity range to 0-1 and computed several reduced version of these raw data as depicted in \autoref{fig:img_dm_reduction}.  

\begin{figure}[!ht]
\centering
\includegraphics[scale=.043]{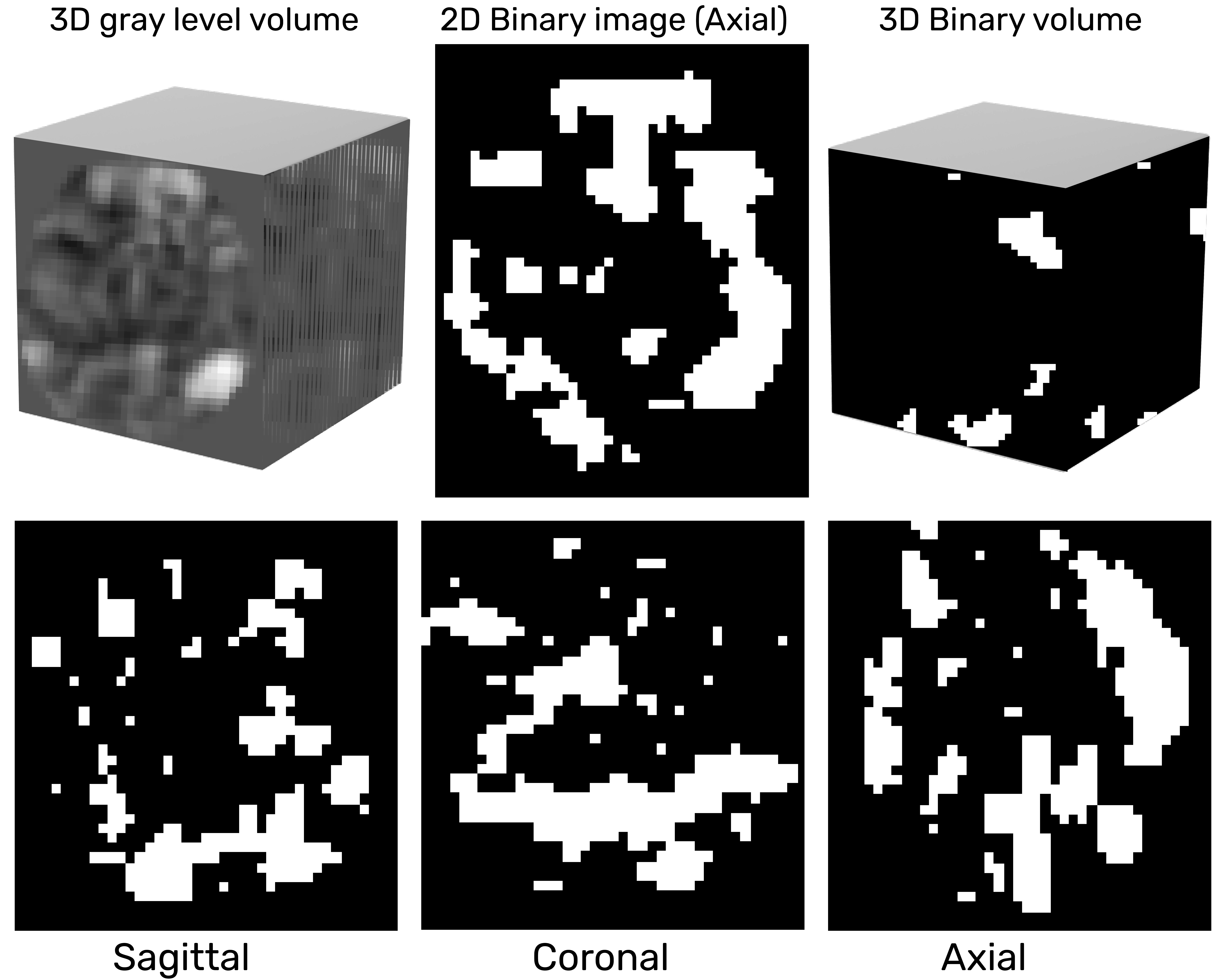}
\caption{fMRI (LANG network) image spatial transformation.}
 \label{fig:seg}
\end{figure}

First, one can reduce the number of spatial dimension via a projection. We produced 2D gray level image by performing \setfont{lmtt}{\textit{Mean}} operation on pixel intensity across the axial (A) plane as shown in \autoref{fig:img_dm_reduction} 
Secondly, to understand whether the intensity of the activation map holds discriminative information, we created 2D binary images by performing an \setfont{lmtt}{\textit{OR}} operation in respect to sagittal, coronal and axial (SCA) plane respectively, which were further stacked together to provide SCA binary stack image. Also, we performed another \setfont{lmtt}{\textit{OR}} operation across the axial plane to obtain a 3D binary volume image which overall, resulted in 4 variants of generated images as illustrated in \autoref{fig:img_dm_reduction}.
Lastly, we tested if the full resolution of voxels is necessary for the classification of the functional network, which are rather formed by large structures than fine details. To this purpose, segmentation of the gray level activation map was performed using SLIC algorithm \cite{achanta2012slic, bakkari2016features}. We processed the 2D segmented labels to obtain a superpixels image, while the 3D segmented labels provided the supervoxels image as shown in \autoref{fig:seg}. Furthermore, we averaged (smoothened) the pixel intensities within each segment of our superpixels and supervoxels images. This step allows us to evaluate the integrity of the functional brain network features which was done by training a CNN model for 7 distinct functional brain network classification using the generated superpixels/supervoxels images.

\begin{figure}[!ht]
\centering
\includegraphics[scale=.06]{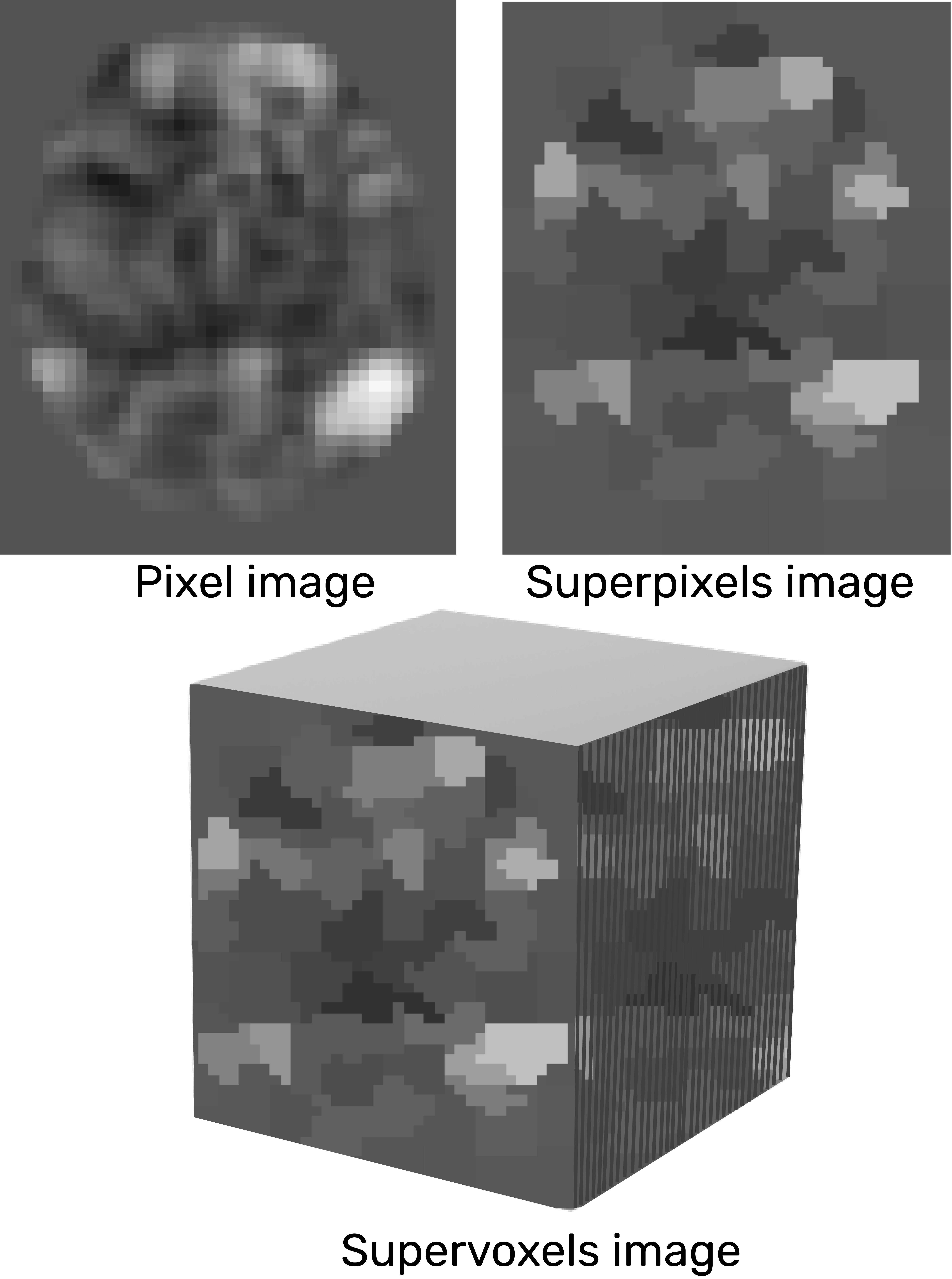}
\caption{Pixel image segmentation into superpixels and supervoxels.}
\label{fig:planar}
\end{figure}

\begin{table}[!ht]
\centering
{\fontsize{7}{9}\selectfont 
\caption{CNN based fMRI brain network classification with unhealthy data.}
\begin{tabular}{|l|c|c|c|} 
\hline
\multicolumn{1}{|c|}{\textbf{Data}}    & \textbf{Train-Test} & \textbf{Accuracy} & \textbf{Parameters}  \\ 
\hline
3D gray level                          & 315-70              & 0.75 ± 0.01       & 2,356,807             \\ 
\hline
3D binary                              & 315-70              & 0.66 ± 0.02       & 2,356,807             \\ 
\hline
2D gray level                          & 315-70              & 0.68 ± 0.01       & 2,337,799             \\ 
\hline
2D binary                              & 315-70              & 0.63 ± 0.01       & 2,337,799             \\ 
\hline
\multicolumn{1}{|c|}{SCA-binary stack} & 315-70              & 0.68 ± 0.02       & 2,011,271             \\
\hline
\end{tabular}
\label{table:gnn_result_cnn} }
\end{table}

When using the dimension reduction from 3D to 2D or from grey level to binary images, we observe performance drop as provided in \autoref{table:gnn_result_cnn}. This suggests that, there is information in the gray level distribution and the 3D shape of the network which are not preserved via the simple spatial dimension reduction tested. By contrast, the values in \autoref{table:cnn_pixel_voxel} represent the functional brain network classification results with CNN model using pixels, superpixels and supervoxels data respectively. Interestingly the loss of performance is very limited when one reduces the gray levels to the average value of the pixels inside a supervoxel or even a superpixel image. Therefore, despite the spatial dimension reduction tested, the reduction of the number of parameters in the models is so far very limited or negligible. To produce this reduction of the model, we proposed to encode the most promising dimension reduction technique (supervoxels) in a compact way as described in the next section.

\begin{table}[!ht]
{\fontsize{8}{10}\selectfont 
\caption{CNN classification of functional brain networks using superpixels/supervoxels images generated in the segmentation stage of graph encoding process with unhealthy subjects as input data.}
\centering
\begin{tabular}{|l|c|c|c|} 
\hline
\multicolumn{1}{|c|}{\textbf{Data}} & \textbf{Train-Test} & \textbf{Accuracy} & \textbf{Parameters}  \\ 
\hline
3D gray level          & 315-70         & 0.75 ± 0.01       & 2,356,807            \\ 
\hline
Superpixels image        & 315-70        & 0.69 ± 0.02      & 2,356,807            \\ 
\hline
Supervoxels image        & 315-70       &  0.73 ± 0.02      & 2,356,807            \\
\hline
\end{tabular}
\label{table:cnn_pixel_voxel} }
\end{table}


\section{Graph encoding}
To further benefit from the spatial dimension reduction of the previous section, we investigate the possibility to reduce the complexity of the associated neural networks models with limited reduction of performance on the functional cerebral network recognition. To this purpose, we consider to encode our supervoxelized images into graphs.  Commonly in graphs, interacting nodes are connected by edges whose weights can be defined by either temporal connections or anatomical junctions, because, graphs are naturally good at relational organization between entities, which makes them great option for representing the 3D capture of voxelwise signals mapped to a specific region of the brain \cite{ahmedt2021graph}. Therefore, a possibly efficient representation of these fMRI network activations in images can be tested using a graph relation network, which connects nodes of related regions via graph edges.



\begin{figure}[!ht]
\centering
\includegraphics[scale=.045]{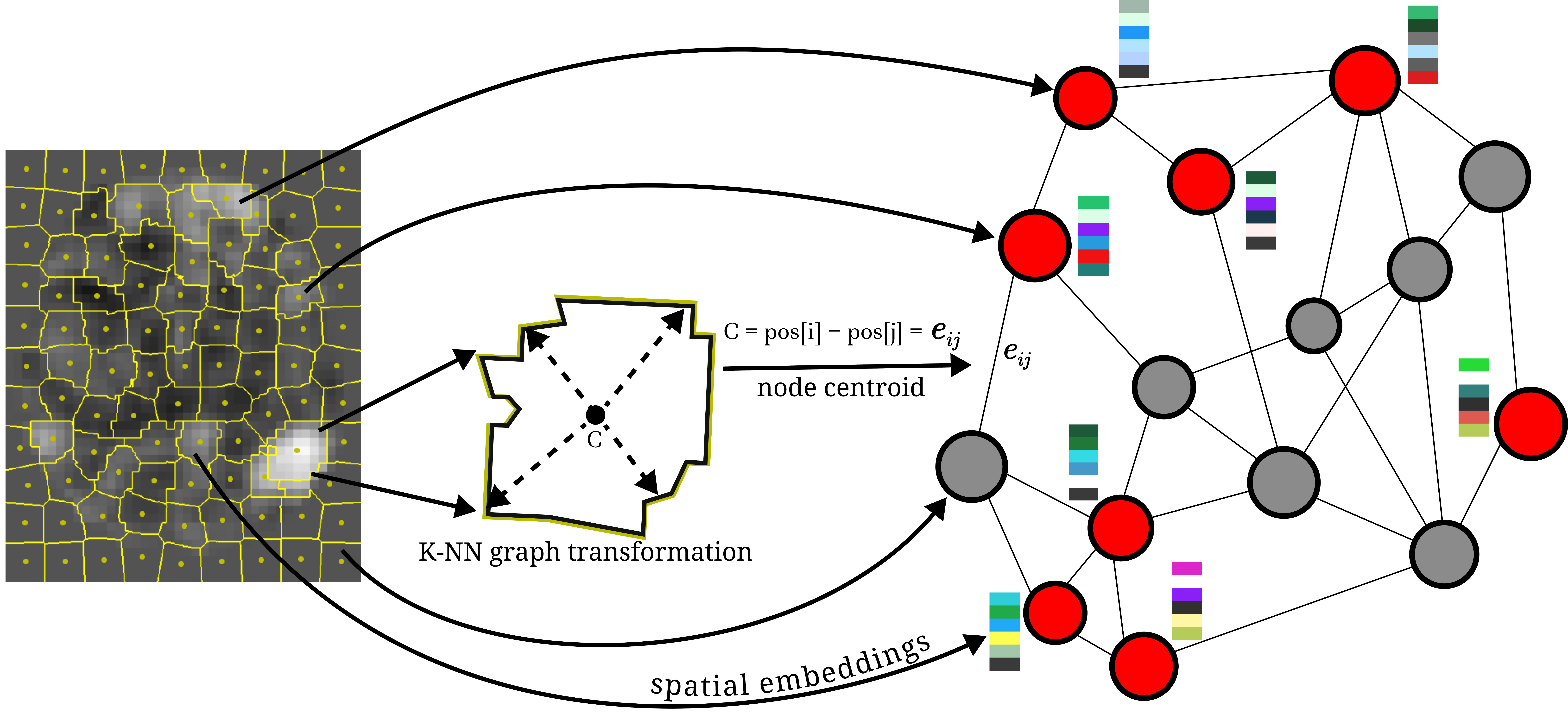}
\caption{Graph encoding process from superpixels/supervoxels images.} 
\label{fig:encoding}
\end{figure}

To obtain a graph representation of our supervoxels images, we connected the segmented neighboring regions through an edge, and denoted the center of each region as a graph node, segment-wise attributes were encoded as node spatial embeddings. This step was repeated until all neighboring nodes were traversed (see \autoref{fig:encoding}). \color{black}{We implemented this approach using the region adjacency graph technique \cite{hagberg2008exploring}, which simply represents each region of the segment as graph nodes and the link between two touching regions as edge using the provided labels of the segmented regions \cite{wu2022graph}. From the extracted relative spatial coordinates of each superpixel of our image data via the cartesian function, we computed the node position as edge attribute ($pos[i] - pos[j]$) via k-NN graph transformation. }
\color{black}{}



The number of supervoxels was fixed empirically based on the typical size of the activation spots.
The resulting graphs from the encoding stage were observed to be structurally indistinguishable from the connectivity point of view. The contrastive information is expected to stand on the distribution of edge values, which differ from one structural network map to another.

 We implemented our method using SplineCNN, a graph neural network which uses a novel type of spline-based convolutional layer for learning \cite{fey2019fast}. This state-of-the-art GNN is suitable for image-based graph classification task because, it allows the capture of local patterns using spatial relationship between graph nodes by performing global graph pooling. We trained our model parameters with 2 convolutional layers and 2 fully connected output layers with indication of 7 classes in the output layer and a softmax activation. Best results were obtained by training with 2-step learning rate values of $1e-3$ for epochs $0-200$ and $1e-5$ for epochs $200-500$ with early stopping.

For fair comparison with the best result obtained with CNN model in \cite{ismaila2022transfer}, \color{black}{}we performed transfer learning during the training of the CNN and GNN models using 80\% - 10\% - 10\% ratio for train-validation-test data slit respectively, as well as early stopper with patient set to 10 misses. \color{black}{} The performance provided in \autoref{tab:transfer} shows the recorded result from fMRI functional network classification using this transfer learning strategy. Brute transfer indicates the strategy of training directly on healthy data and testing on unhealthy data for both CNN and GNN models. In this cohort, results were compared with values from training and testing on unhealthy data using CNN and GNN model, which provided the $1^{st}$ baseline and $2^{nd}$ baseline values of 0.75 ± 0.01 and  0.64 ± 0.03 respectively, while 0.78 ± 0.01 and  0.70 ± 0.01 were recorded in the transfer learning approach with CNN and GNN respectively. As a consequence, we demonstrate the possibility to obtain a compression of a factor of 26 on the number of model parameters after supervoxeization and graph encoding with only a reduction of $8\%$.



%

\begin{table}[!ht]
\centering
{\fontsize{7}{9}\selectfont 
\caption{Transfer learning classification with CNN and GNN models.}
\begin{tabular}{|l|l|c|c|} 
\hline
\multicolumn{1}{|c|}{\textbf{Data}}                                     & \multicolumn{1}{c|}{\textbf{Train-Test}} & \textbf{Accuracy} & \textbf{Parameters}  \\ 
\hline
CNN brute-transfer                                                      & healthy-unhealthy       & 0.75 ± 0.01       & 2,356,807             \\ 
\hline
\begin{tabular}[c]{@{}l@{}}CNN fMRI healthy \\(pretrained)\end{tabular} & unhealthy-unhealthy      & 0.78 ± 0.01       & 2,356,807             \\ 
\hline
GNN brute-transfer                                                      & healthy-unhealthy       & 0.67 ± 0.02       & 90,855               \\ 
\hline
\begin{tabular}[c]{@{}l@{}}GNN fMRI healthy \\(pretrained)\end{tabular} & unhealthy-unhealthy      & 0.70 ± 0.01       & 90,855               \\ 
\hline
\end{tabular}
\label{tab:transfer} }
\end{table}

\section{Conclusion}

In this study, we investigated ways to reduce the complexity of end-to-end machine learning models based on convolutional neural networks for the automatic recognition of functional cerebral networks via resting-state fMRI data.
A compaction of the activation maps into superpixels or supervoxels shows limited impact on the classification performance. \color{black}{} We emphasize the anticipated influence of our 3D multi-channel images in model parameters, which motivates exploration of a dimension reduction technique before introducing the graph encoding technique. Model evaluation based on spatial dimension reduction was done to investigate its minimal influence in reducing our model parameter. However, this stage was important towards more efficient data encoding (graph structure), which was later shown to have significantly reduced the model parameter.\color{black}{} Our initial encoding effort produces a compression of a factor $26\times$ where associated reduction in performance was observed at only $8\%$. 

The effort to reduce the complexity of the models was concentrated on the encoding approach of our fMRI data. It would naturally be interesting to couple such effort with investigation on the architecture of the models \cite{sanchez2020tinyml, guo2022machine}.

\bibliographystyle{IEEEtran}
\bibliography{reference}

\end{document}